\documentclass[10pt,journal,compsoc]{IEEEtran}

\usepackage{booktabs} 
\usepackage{amsthm}
\theoremstyle{plain}
\usepackage[compress]{cite}
\usepackage{graphics}
\usepackage{epstopdf}
\usepackage{epsfig}
\usepackage{lineno,hyperref}
\usepackage{array}
\usepackage{amsmath,amssymb}
\usepackage{mathrsfs}
\usepackage{subfigure} 
\usepackage{multirow}
\usepackage{xcolor}
\usepackage{bm}
\usepackage{eqparbox}
\usepackage{algorithm}  
\usepackage{algorithmic}
\usepackage{graphicx}

\usepackage[justification=centering]{caption}
\usepackage{amsfonts}
\usepackage{enumerate}
\usepackage{indentfirst}
\usepackage{CJK}
\usepackage{url}

\usepackage{ragged2e}

\begin{document}
\title{Unsupervised Graph Embedding via Adaptive Graph Learning}	
% Single author syntax
\author{
	Rui Zhang,\IEEEmembership{~Member,~IEEE}, Yunxing Zhang, Xuelong Li$^*$,\IEEEmembership{~Fellow,~IEEE}
	\thanks{Xuelong Li$^*$ is the corresponding author.}
	
	% <-this % stops a space
	\thanks{Rui Zhang, Yunxing Zhang, and Xuelong Li are with the School of Computer Science and Center for OPTical IMagery Analysis and Learning (OPTIMAL), Northwestern Polytechnical University, Xi'an 710072, Shaanxi, P. R. China.}
	
	\thanks{E-mail: ruizhang8633@gmail.com, zhangyunxing423@outlook.com, and xuelong\_li@nwpu.edu.cn }
	
}
\markboth{IEEE TRANSACTIONS ON PATTERN ANALYSIS AND MACHINE INTELLIGENCE}%
{Zhang \MakeLowercase{\textit{et al.}}: Unsupervised Graph Embedding via Adaptive Graph Learning}

\IEEEtitleabstractindextext{
\begin{abstract}
	
\justifying Graph autoencoders (GAEs) are powerful tools in representation learning for graph embedding.
However, the performance of GAEs is very dependent on the quality of the graph structure, i.e., of the adjacency matrix.
In other words, GAEs would perform poorly when the adjacency matrix is incomplete or be disturbed.
In this paper, two novel unsupervised graph embedding methods, \emph{unsupervised graph embedding via adaptive graph learning} (BAGE) and \emph{unsupervised graph embedding via variational adaptive graph learning} (VBAGE) are proposed.
The proposed methods expand the application range of GAEs on graph embedding, i.e, on the general datasets without graph structure.
Meanwhile, the adaptive learning mechanism can initialize the adjacency matrix without be affected by the parameter.
Besides that, the latent representations are embedded in the laplacian graph structure to preserve the topology structure of the graph in the vector space.
Moreover, the adjacency matrix can be self-learned for better embedding performance when the original graph structure is incomplete.
With adaptive learning, the proposed method is much more robust to the graph structure.
Experimental studies on several datasets validate our design and demonstrate that our methods outperform baselines by a wide margin in node clustering, node classification, and graph visualization tasks.

\end{abstract}
}

\maketitle

\IEEEdisplaynontitleabstractindextext

\IEEEpeerreviewmaketitle

\IEEEraisesectionheading{\section{Introduction}}
\IEEEPARstart{G}raphs are powerful tools to seek the geometric structure of data and graph analysis has been attracting increasing attention in recent years due to the ubiquity of networks in the real world.
There are various applications using graphs in machine learning and data mining fields such as node classification \cite{wang2020second} \cite{bai2020learning}, node clustering \cite{wang2020large} \cite{li2020multi}, link prediction \cite{shao2019community}, and visualization \cite{gao2020learning}.
However, it is difficult to directly apply the existing machine learning methods to graph data, due to the high computational complexity, low parallelizability, and inapplicability of most methods to graph data \cite{cui2018survey}.

In response to these problems, many graph embedding methods have been proposed in recent years.
As one of the representation learning, the purpose of graph embedding is to learn the low-dimensional feature vectors which should preserve the topology structure of the graph.
In the early 2000s, researchers developed graph embedding algorithms as part of dimensionality reduction techniques.
The early graph embedding methods such as Laplacian Eigenmaps \cite{belkin2002laplacian} and Locally Linear Embedding (LLE) \cite{roweis2000nonlinear} map the nodes of the graph into a low-dimensional vector space.
Since 2010, research on graph embedding has shifted to obtaining scalable graph embedding techniques that leverage the sparsity of real-world networks \cite{goyal2018graph}.
The methods at this stage such as Graph Factorization \cite{ahmed2013distributed}, LINE \cite{tang2015line}, HOPE \cite{ou2016asymmetric}, and SDNE \cite{wang2016structural} attempt to preserve both first order and second proximities.

In recent years, graph neural networks (GNNs) \cite{kipf2016semi} emerge as powerful node embedding methods with successful applications in broad areas such as social networks, recommended systems, and natural language processing.
Graph neural networks (GNNs) are powerful tools in representation learning for graphs and able to incorporate sparse and discrete dependency structures between data points.
Graph neural networks (GNNs) could be roughly categorized into four categories: recurrent graph neural networks (RecGNNs), convolutional graph neural networks (ConvGNNs), graph autoencoders (GAEs), and spatial-temporal graph neural networks (STGNNs) \cite{wu2020comprehensive}.

Graph autoencoders (GAEs) are the effective unsupervised learning frameworks that encode the node features and graph construction into the latent representations and decode the graph construction. 
GAEs and most of their extensions rely on graph convolutional networks (GCN) to learn vector space representations of nodes.
GAEs can be used to learn graph (network) embeddings \cite{pan2018adversarially} and graph generative distributions \cite{li2018learning}. 
For graph embedding, GAEs mainly learn the latent representations by reconstructing the graph construction, e.g., the adjacency matrix. 
For graph generation, GAEs can learn the generative distribution of graphs and are mostly designed to solve the molecular graph generation problem \cite{li2018learning}.
The main distinction between GAEs and graph embedding is that GAEs are designed for various tasks while graph embedding covers various kinds of methods targeting the same task. 

Although GNNs achieve great success from in representation learning of graphs, recent studies show that the performance of GNNs is very dependent on the quality of the adjacency matrix.
In other words, GNNs will perform poorly while the adjacency matrix is under attack or incomplete.
The incomplete means the adjacency matrix is partially missing or be disturbed.
To defend against adversarial attacks, Jin et al. \cite{jin2020graph} propose a framework Pro-GNN, which can jointly learn a structural graph and a robust graph neural network model from the perturbed graph guided by these properties.
For the incomplete, Chen et al. \cite{chen2019deep} propose a graph learning framework that jointly learning the graph structure and graph embeddings simultaneously.

However, these methods are all designed for the supervised graph neural networks (GNNs) not the unsupervised graph autoencoder (GAEs).
Besides that, they use $k$-nearest neighbor ($k$NN) to initialize the adjacency matrix when the graph structure is unavailable.
A major shortcoming of this approach is that the efficacy of the resulting models hinges on the choice of $k$ \cite{franceschi2019learning}.
In any case, the graph creation and parameter learning steps are independent and require heuristics and trial and error.

In this paper, two novel unsupervised graph embedding methods, \emph{unsupervised graph embedding via adaptive graph learning} (BAGE) and \emph{unsupervised graph embedding via variational adaptive graph learning} (VBAGE) are developed.
The contributions can be summarized below:

$\bullet$ The proposed method expands the application range of GAEs on graph embedding, i.e, on the general datasets without graph structure.
The adaptive learning mechanism is able to initialize the adjacency matrix without be affected by the parameter $k$.

$\bullet$ The latent representations are embedded in the laplacian graph structure to preserve the topology structure of the graph in the vector space.

$\bullet$ With adaptive learning, the adjacency matrix can be self-learned for better embedding performance which enhances the robustness of the model.

\section{Related Work}
In this section, we outline the background and development of graph embedding and graph autoencoders (GAEs).

The goal of graph embedding is to learn the low-dimensional latent representations of nodes that preserve the topological information of the graph.
The early methods such as DeepWalk \cite{perozzi2014deepwalk} uses a random walk to generate sequences of nodes from a network and transform graph construction information into linear sequences.
Inspired by DeepWalk, DRNE \cite{tu2018deep} adopts a Long Short Term Memory (LSTM) network to aggregate a node's neighbors. 
Similar to DRNE, NetRA \cite{yu2018learning} also uses the LSTM network with random walks rooted on each node and regularizes the learned network embeddings within a prior distribution via adversarial training.
The embedding method SDNE \cite{wang2016structural} exploits the first-order and second-order proximity jointly to preserve the network structure.
The structure of SDNE is very similar to the graph autoencoders and can be seen as an early approach of GAEs.

The relationship between GAEs and graph embedding can be understood as: GAEs are a general term for a series of methods, and graph embedding is one of the tasks that GAE can perform.
Earlier GAE approaches such as DNGR \cite{cao2016deep} and SDNE \cite{wang2016structural} mainly build the GAE frameworks for graph embedding by multi-layer perceptrons. 
Nevertheless, DNGR and SDNE only consider node structure information but ignore the node features information. 
In other words, the early GAE methods directly learn the node embeddings from a graph where each node on this graph does not contain feature information.
What really kicked off graph autoencoder is \cite{kipf2016variational} whose methods GAE and VGAE laid the foundation for the later GAE methods.
Inspired by generative adversarial networks (GANs) and VGAE \cite{kipf2016variational}, Adversarially Regularized Variational Graph Autoencoder (ARGE and ARVGE) \cite{panadversarially} is proposed that endeavors to learn an encoder that produces the empirical distribution.
By replacing the GCN encoder with a simple linear model, Salha et al. \cite{salha2019keep} propose a linear graph autoencoder (LGAE and LVGAE) which is more simple.

Our methods (BAGE and VBAGE) also use the graph autoencoder to learn graph embedding.
However, our method differs from these methods in the following points:

\noindent 1) Our methods can be applied to more general datasets, i.e., the dataset without the graph structure.\\
2) The learned latent representations are embedded into the laplacian graph structure to preserve the topology structure of the graph in the vector space.\\
3) The adjacency matrix in our framework can be adaptively learned, which enhances the robustness of the model.

We will detailly describe these aspects in the rest of the paper.

\section{Notations and Problem Statement}
Before we present the problem statement, we first introduce some notations and basic concepts.
The Frobenius norm of a matrix $\mathbf{A}$ is defined by $\|\mathrm{A}\|_{F}^{2} = \Sigma_{i j} a_{i j}^{2}$.
We use $\odot$ to denote the Hadamard product of matrices and $tr(A)$ to indicate the trace of matrix $\mathbf{A}$, i.e., $t r(\mathbf{A}) = \sum_{i} a_{i i}$.
Epoch in the paper means the number of iteration.

Let $\mathbf{G = \{V,E,X\}}$ be a graph, where $\mathbf{V} = \{v_1,v_2, \dots, v_n\}$ is a set of nodes with $|\mathbf{V}| = n$ and $\mathbf{E}$ is the set of connecting edges among each node.
The edges describe the relations between nodes and can also be represented by an adjacency matrix $\mathbf{A} = [a_{ij}] \in \mathbb{R}^{n \times n}$ where $a_{ij}$ denotes the relation between nodes $v_i$ and $v_j$.
Furthermore, we use $\mathbf{X}=\left[\mathbf{x}_{1}, \mathbf{x}_{2}, \cdots, \mathbf{x}_{n}\right]^{T} \in \mathbb{R}^{n \times m}$ to denote the node feature matrix where $\mathbf{x}_i$ is the feature vector of node $v_i$. 
The number of nodes is $n$ and $m$ is the dimension of the raw data. 

The aim of graph embedding in our method is to learn the low-dimensional latent representation $\mathbf{Z} \in \mathbb{R}^{n \times f}$ from the node matrix $\mathbf{V}$ with the formal format as: $f:(\mathbf{A}, \mathbf{X}) \mapsto \mathbf{Z}$.
The learned latent representation $\mathbf{Z}$ in latent space should preserve the topological structure of the graph as well as node feature information.

\section{Framework}

\begin{figure*}[h]
	\centering	
	\includegraphics[height=8cm, width=16cm]{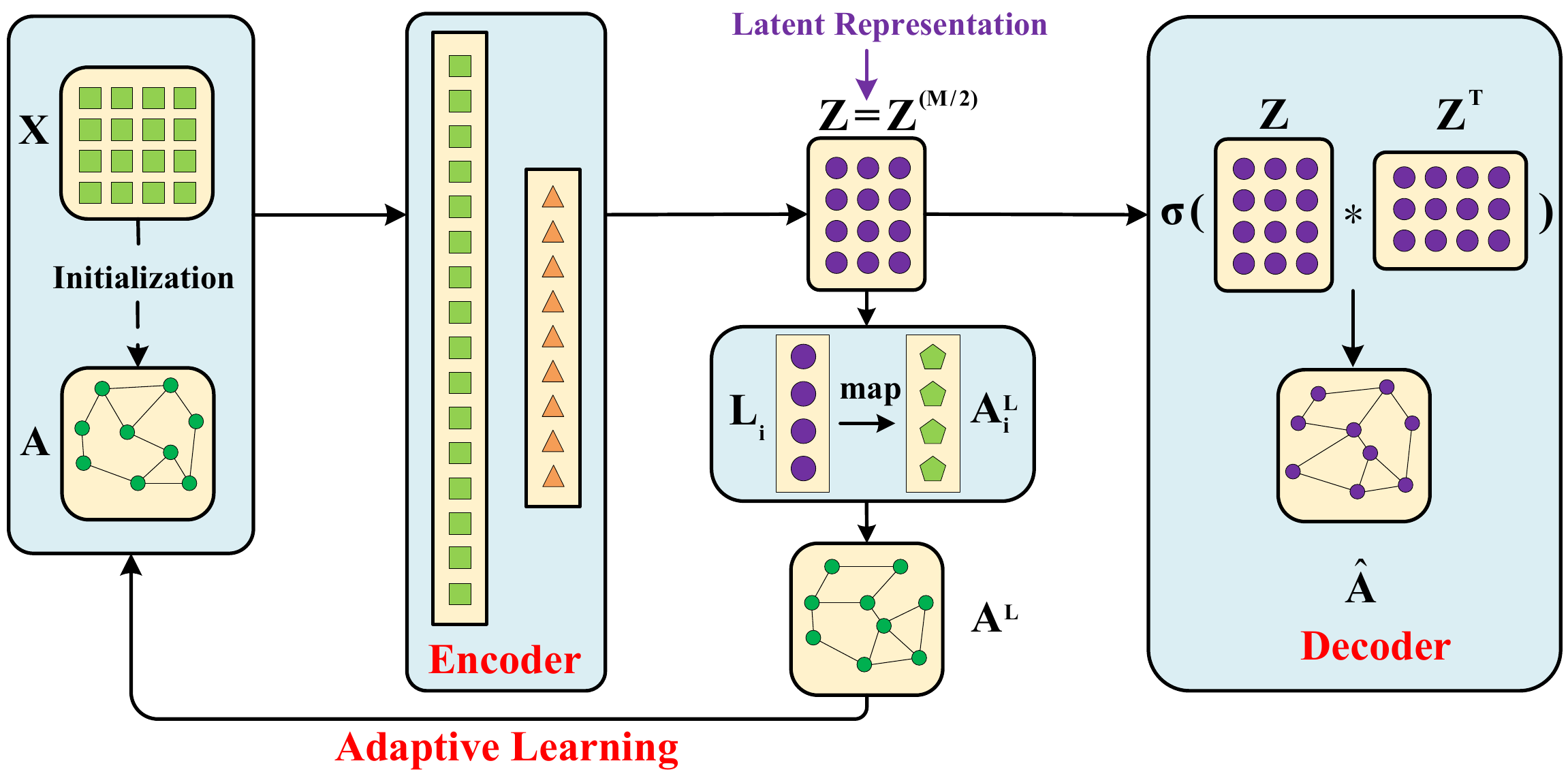}
	\caption{The architecture of the proposed framework (See Algorithm \ref{BAGE} for details).}

	\label{Flow_Block}
\end{figure*} 

Fig. \ref{Flow_Block} illustrates the workflow of our methods (\emph{BAGE} and \emph{VBAGE}) that consists of three modules: the graph convolutional encoder, the decoder, and the laplacian graph structure.

\subsection{Graph Convolutional Encoder Model} 

The encoder model learns a layer-wise transformation by a spectral graph convolutional function $f\left(\mathbf{Z}^{(l)}, \mathbf{A} | \mathbf{W}^{(l)}\right):$
\begin{equation} \label{Z_1}
\mathbf{Z}^{(l+1)} = f(\mathbf{Z}^{(l)}, \mathbf{A}|\mathbf{W}^{(l)}),
\end{equation}
where $\mathbf{Z}^{(l)}$ is the input for convolution and $\mathbf{Z}^{(l+1)}$ is the output after convolution.
$\mathbf{W}^{(l)}$ is the weight parameter matrix that needs to be learned in the neural network. 
In this paper, $\mathbf{Z}^{(0)} = \mathbf{X} \in \mathbb{R}^{n \times m}$  is the input node features matrix. 
Specifically speaking, each layer of our graph convolutional network can be calculated as follows:
\begin{equation} \label{Z_2}
f\left(\mathbf{Z}^{(l)}, \mathbf{A} | \mathbf{W}^{(l)}\right)=\phi\left(\widetilde{\mathbf{D}}^{-\frac{1}{2}} \widetilde{\mathbf{A}} \widetilde{\mathbf{D}}^{-\frac{1}{2}} \mathbf{Z}^{(l)} \mathbf{W}^{(l)}\right).
\end{equation}
Here, $\widetilde{\mathbf{A}} = \mathbf{A} + \mathbf{I}$, $\widetilde{\mathbf{D}}_{ii} = \sum_j \widetilde{\mathbf{A}}_{ij}$, $\mathbf{I}$ is the identity matrix of $\mathbf{A}$, and $\phi$ is the activation function such as Relu($t$) = $\max(0, t)$. 

$\bullet$ The \emph{Graph Encoder} in our method (BAGE) is like \cite{kipf2016variational} and \cite{pan2018adversarially}, which is constructed as follows:
\begin{equation} \label{relu}
\mathbf{Z}^{(1)} =f_{\text {Relu}}(\mathbf{X}, \mathbf{A} | \mathbf{W}^{(0)});
\end{equation}
\begin{equation} \label{linear}
\mathbf{Z}^{(2)} =f_{\text {linear }}(\mathbf{Z}^{(1)}, \mathbf{A} | \mathbf{W}^{(1)}).
\end{equation}

$\bullet$ The \emph{Variational Graph Encoder} in our method (VBAGE) is defined by an inference model:
\begin{equation}
\begin{aligned}
q(\mathbf{Z} \mid \mathbf{X}, \mathbf{A}) & = \prod_{i=1}^{n} q\left(\mathbf{z}_{\mathbf{i}} \mid \mathbf{X}, \mathbf{A}\right), \\
q\left(\mathbf{z}_{\mathbf{i}} \mid \mathbf{X}, \mathbf{A}\right) &= \mathcal{N}\left(\mathbf{z}_{i} \mid \boldsymbol{\mu}_{i}, \operatorname{diag}\left(\boldsymbol{\sigma}^{2}\right)\right).
\end{aligned}
\end{equation}

Here, $\boldsymbol{\mu}=\mathbf{Z}^{(2)}$ is the matrix of mean vectors $\mathbf{z}_{i} ;$ similarly $\log \boldsymbol{\sigma}=f_{\text {linear }}\left(\mathbf{Z}^{(1)}, \mathbf{A} \mid \mathbf{W}^{\prime(1)}\right)$ which share the weights $\mathbf{W}^{(0)}$
with $\boldsymbol{\mu}$ in the first layer in Eq. (\ref{relu}).

\subsection{Decoder Model} 

The decoder model reconstructs the graph from the learned latent representations and the reconstructed adjacency matrix $\mathbf{\widehat{A}}$ can be represented as follows:
\begin{equation}
\mathbf{\widehat{A}} = {\rm sigmoid} (\mathbf{Z}\mathbf{Z}^\emph{T}),
\end{equation}
where $\mathbf{Z}$ is the latent representation and $\mathbf{Z} = \emph {encoder} \ (\mathbf{Z}|\mathbf{X,A})$.

In terms of loss function, we did not use the cross-entropy loss function to define the reconstruction loss like \cite{kipf2016variational} and \cite{pan2018adversarially}.
We impose more penalty on the reconstruction error of the non-zero elements than that of zero elements.

$\bullet$ The reconstruction loss of the graph construction for BAGE is calculated as:
\begin{equation} \label{graph_loss_BAGE}
\begin{aligned}
\mathcal{L}_{G1} &=\sum_{i=1}^{n}\left\|\left(\mathbf{a}_{i}-\hat{\mathbf{a}}_{i}\right) \odot \mathbf{b}_{\mathbf{i}}\right\|_{2}^{2},\\
&= \|(\mathbf{A} - \mathbf{\widehat{A}}) \odot \mathbf{B}\|_F^2,
\end{aligned}
\end{equation}
where $\odot$ means the Hadamard product, $\mathbf{b}_{\mathbf{i}}=\left\{b_{i, j}\right\}_{j=1}^{n} .$ If $a_{i, j}=$ $0, b_{i, j}=1,$ else $b_{i, j}=\beta>1.$ 

$\bullet$ The reconstruction loss of the graph construction for VBAGE is calculated as:
\begin{equation} \label{graph_loss_VBAGE}
\mathcal{L}_{G2}=\|(\mathbf{A} - \mathbf{\widehat{A}}) \odot \mathbf{B}\|_F^2 -\mathbf{KL}[q(\mathbf{Z} \mid \mathbf{X}, \mathbf{A}) \| p(\mathbf{Z})],
\end{equation}
where $\mathbf{K L}[q(\boldsymbol{\bullet}) \| p(\boldsymbol{\bullet})]$ is the Kullback-Leibler divergence between $q(\bullet)$ and $p(\bullet) .$ We also take a Gaussian prior $p(\mathbf{Z})=$ $\prod_{i} p\left(\mathbf{z}_{i}\right)=\prod_{i} \mathcal{N}\left(\mathbf{z}_{i} \mid 0, \mathbf{I}\right)$ like \cite{pan2018adversarially} and \cite{kipf2016variational}.

\subsection{Laplacian Graph Structure} 

The reconstruction loss in Eqs. (\ref{graph_loss_BAGE}) and (\ref{graph_loss_VBAGE}) is only focused on graph reconstruction while ignoring latent representations.
Aim at this, we embed the latent representations into the laplacian graph structure.
The loss function for this goal is defined as follows: 

\begin{equation} \label{Laplacian}
\begin{aligned}
\mathcal{L}_{L} &= \sum_{i,j=1}^{n}\left(\left\|\mathbf{z}_{i}-\mathbf{z}_{j}\right\|_{2}^{2} a_{i j} + \gamma_i a_{i j}^{2}\right)\\
&= {\rm tr}(\mathbf{ZLZ}^T)+{\gamma} \mathbf{\|A\|}_{F}^2\\	
&\text {s.t.} \quad \mathbf{a}_{i}^{T} \mathbf{1}=1, \mathbf{0} \leq \mathbf{a}_{i} \leq \mathbf{1},
\end{aligned}	
\end{equation}
where $\gamma$ is a regularization parameter that can be adaptively solved.
The laplacian graph structure is also the basis of the adaptive learning of the adjacency matrix that is also the biggest contribution of this paper.

The objective function of Eq. (\ref{Laplacian}) borrows the idea of Laplacian Eigenmaps \cite{belkin2003laplacian}, which incurs a penalty when similar latent representations are far away in the embedding space.
The Laplacian matrix $\mathbf{L}$ is defined as:
\begin{equation}
\mathbf{L = D -A},
\end{equation}
where the $\mathbf{D}$ is the degree matrix whose diagonal element $d_{ii} = \sum_{j} a_{ij}$.

Similar ideas appear in some works on graph learning and network embedding.
The difference between our methods and them is that we directly incorporate the laplacian graph structure into the latent representations, not into the feature matrix, like CAN \cite{nie2014clustering} or into the label matrix, like SDNE \cite{wang2016structural}.
Thus laplacian graph structure can make the vertexes linked by an edge be mapped near in the embedding space.

$\bullet$ In summary, the loss function of the BAGE can be written as:
\begin{equation} \label{loss_BAGE}
\mathcal{L}_{BAGE} = \mathcal{L}_{G1} + \lambda \mathcal{L}_L  + \nu \mathcal{L}_{reg}.
\end{equation}

$\bullet$ And the loss function of the VBAGE can be written as:
\begin{equation} \label{loss_VBAGE}
\mathcal{L}_{VBAGE} = \mathcal{L}_{G2} + \lambda \mathcal{L}_L  + \nu \mathcal{L}_{reg},
\end{equation}
where $\mathcal{L}_{reg}$ is the $\ell_2$-norm regularizer term with coefficient $\nu$ to prevent overfitting, which is defined as follows:
\begin{equation}
\mathcal{L}_{reg} = \frac{1}{2} \sum_{i} (\|\mathbf{W}^{(i)} \|_F^2).
\end{equation}

\section{Adaptive Learning of the Adjacency Matrix}

The most important contribution of this paper is the adaptive learning of the adjacency matrix that tries to answer the following two questions:

\noindent$\bullet$ When the original graph structure is incomplete,  can we learn an alternative graph structure to obtain better embedding effects?

\noindent$\bullet$ When we apply graph convolution to the data without an initial graph structure, can we get the graph structure through adaptive learning rather than $k$NN initialization?

\subsection {The Solution of The Adjacency Matrix}
In practical applications, the adjacency matrix with adjustable sparsity tends to bring better results.
And that is a reason why we do not update the adjacency matrix by back-propagation algorithm directly, which will produce a meaningless dense matrix.  
The adaptive learning of the adjacency matrix is based on the laplacian graph structure in Eq. (\ref{Laplacian}) as follows:
 
\begin{equation}
\begin{aligned}
	&\min_{a_{ij}} \sum_{i,j=1}^{n}\left(\left\|\mathbf{z}_{i}-\mathbf{z}_{j}\right\|_{2}^{2} a_{i j} + {\gamma_i} a_{i j}^{2}\right)\\
	& \text {\ s.t.} \quad \mathbf{a}_{i}^{T} \mathbf{1}=1, \mathbf{0} \leq \mathbf{a}_{i} \leq \mathbf{1}.
\end{aligned}
\label{a_q1}
\end{equation}
Let us denote the distance between two nodes as $h_{ij}$, i.e., $h_{ij} = \left\|\mathbf{z}_{i}-\mathbf{z}_{j}\right\|_{2}^{2}$. 
The $j$-th element of vector $\mathbf{h}_i \in \mathbb{R}^{n \times 1}$ is $j$-th element $h_{ij}$ and $\mathbf{a}_i \in \mathbb{R}^{n \times 1}$ is a vector with its $j$-th element $a_{ij}$. 
The Lagrange equation of problem (\ref{a_q1}) is represented as:
\begin{equation} \label{a_L}
\mathcal{L}(\mathbf{a}_i, \eta, \bm{\zeta}_{i})= \frac{1}{2}\left\|\mathbf{a}_{i}+\frac{\mathbf{h}_{i}}{2 \gamma_{i}}\right\|_{2}^{2}-\eta(\mathbf{a}_i^T \mathbf{1}-1)-\bm{\zeta}_{i}^{T} \mathbf{a}_i,
\end{equation}
where $\eta$ and $\bm{\zeta}_{i} \geq \mathbf{0}$ are the Lagrange multipliers. 
Using the Karush-Kuhn-Tucker (KKT) conditions, we can derive the optimal solution of ${a}_{ij}$ as:
\begin{equation}\label{a_q2}
a_{i j}=\left(-\frac{h_{i j}}{2 \gamma_{i}}+\eta\right)_{+},
\end{equation}
where $(\bullet)_+ = \max(\bullet,0)$. 
To equip the adjacency matrix with adjustable sparsity, we take only $\emph{k}$ nodes points closest to $\mathbf{z}_i$ into consideration and the parameter $k$ is responsible for adjusting the sparsity of the adjacency matrix.
Therefore, $\mathbf{a}_i$ satisfies $a_{i \emph{k}}>0 \geq a_{i,\emph{k}+1}$ as:
\begin{equation} \label{BDS}
\left\{\begin{array}{l}{a_{i \emph{k}}>0 \Rightarrow-\frac{h_{i \emph{k}}}{2 \gamma_{i}}+\eta>0} \\ 
{a_{i, \emph{k}+1} \leq 0 \Rightarrow-\frac{h_{i, \emph{k}+1}}{2 \gamma_{i}}+\eta \leq 0}\end{array}\right..
\end{equation}
According to Eq. (\ref{a_q2}) and the constraint $\mathbf{a}_{i}^{T} \mathbf{1}=1$, we have:
\begin{equation}
\label{eta}
\sum_{j=1}^{k}\left(-\frac{h_{i j}}{2 \gamma_{i}}+\eta\right)=1 
\Longrightarrow \eta=\frac{1}{k}+\frac{1}{2 k \gamma_{i}} \sum_{j=1}^{k} h_{i j}.
\end{equation}

The overall $\gamma$ is set to the mean of $\gamma_i$ and it can be learned adaptively as:
\begin{equation}\label{gamma}
\gamma=\frac{1}{n} \sum \limits_{i=1}^{n}\left(\frac{k}{2} h_{i,k+1}-\frac{1}{2} \sum_{j=1}^{k} h_{ij}\right).
\end{equation}
Without loss of generality, let us suppose $h_{i1}, h_{i2}, \dots, h_{in}$ are ordered from small to large.
Then the adjacency matrix can be solved as:
\begin{equation}\label{a_ij}
	a_{i j}=\left(\frac{h_{i, \emph{k}+1}- h_{i j}}{\emph{k} h_{i, \emph{k}+1}-\sum_{j=1}^{\emph{k}} h_{i j}}\right)_{+}.
\end{equation}
Based on Eq. (\ref{a_ij}), the adjacency matrix $\mathbf{A}$ can be updated to $\mathbf{A^L}$ by the latent representations.
What is more, the adjacency matrix can also be initialized by Eq. (\ref{a_ij}) when there is no initial graph structure.

\subsection{Update The Adjacency Matrix}
Eq. (\ref{a_ij}) provides a solution to the graph structure when the adjacency matrix does not exist or is incomplete. 
During the iterative process, the learned latent representations can be used to calculate the adjacency matrix for the next iteration.
However, it is harmful to discard the initial graph structure totally since the optimal graph structure is potentially a small shift from the initial graph structure \cite{chen2019deep}.

With this assumption, we combine the learned graph structure with the initial graph structured as follows:
\begin{equation} \label{A_alpha}
\mathbf{A} = \alpha \mathbf{A}_{L} + (1- \alpha) \mathbf{A}_{0},
\end{equation}
where $\mathbf{A}_{0}$ is initial adjacency matrix and $\mathbf{A}_{L}$ is the learned adjacency matrix in the iteration.
A hyperparameter $\alpha$ is used to balance the trade-off between the learned graph structure and the initial graph structure. 
Moreover, we set a threshold $\tau$ for stopping updates, i.e., if ($epoch$ $\textgreater$ $\tau$), then the updating stop.

\subsection {Distribution of Parameter $k$}
There is only one parameter $k$ in Eq. (\ref{a_ij}). 
If we set $k$ as a fixed hyperparameter, it will cause all samples to have the same number of neighbors, i.e., the number of non-zero elements in each row of the adjacency matrix is the same.
It will bring the same disadvantage as $k$NN since the number of neighbors for each sample is mostly different in real-world graph data.

Instead of fixing $k$, we make $k$ to be a parameter that can be learned adaptively.
Since the $k$ represents the number of neighbors for each sample, i.e., the number of non-zero elements in each row of the adjacency matrix, we sample $k$ from the normal distribution.
$k$ obeys a normal distribution with a mean $\mu k$ and a variance of 1, i.e., $k \sim \mathcal{N} (uk, 1)$.
The value of $\mu k$ is the number of neighbors for the sample in the former iteration.
Besides that, we set a maximum and minimum limit on $k$ to keep it in a reasonable range.
Specific to Eq. (\ref{a_ij}), the number of neighbors for each sample in the adjacency matrix will be dynamically obtained during the updating.

\section{Optimization}
To optimize the aforementioned models, the goal is to minimize the loss of $\mathcal{L}_{BAGE}$ \ref{loss_BAGE} and $\mathcal{L}_{VBAGE}$ \ref{loss_VBAGE} which are the function of the neural network parameter $\mathbf{W}$. 
The calculation of the partial derivative of $\mathcal{L}_{BAGE}$ \ref{loss_BAGE} and $\mathcal{L}_{VBAGE}$ \ref{loss_VBAGE} are estimated using the back-propagation. 
Furthermore, the proposed framework is optimized by adaptive moment estimation (Adam), where $T$ is the maximum iteration number.

\begin{figure}[h]
	\centering
	\setlength{\belowcaptionskip}{0pt}
	{
	 \includegraphics[height=3cm,width=9cm]{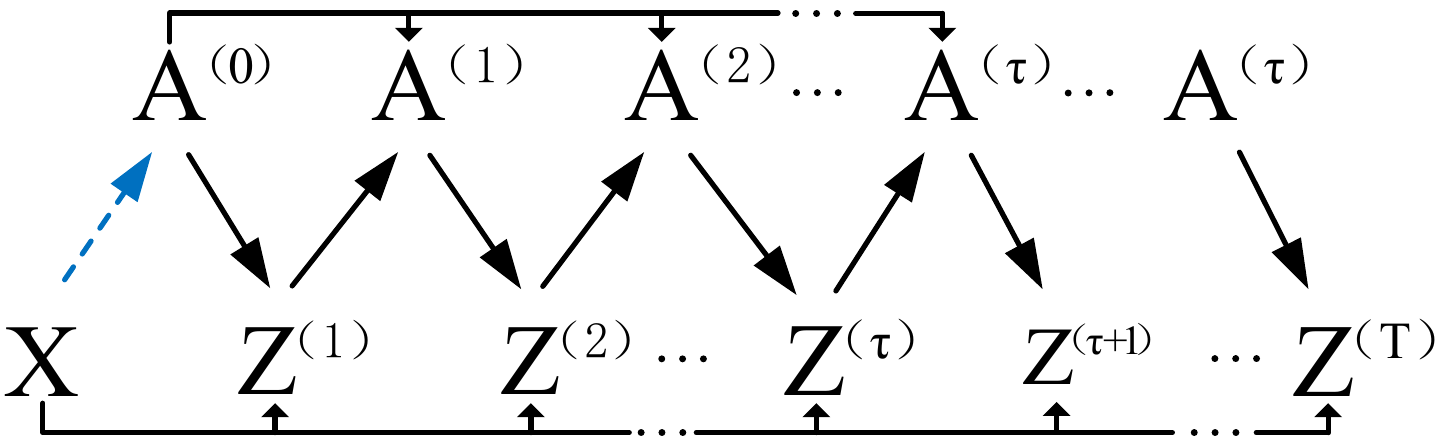}
	}

	\caption{Information flow of the adaptive learning process.}
	\label{Flow_A}
\end{figure}

As for the adaptive learning of the adjacency matrix, the time complexity of the updating process is $\mathcal{O}((d n^{2}) \tau)$.
To show clearly the information transmission in the adaptive learning process, Fig. \ref{Flow_A} is made to show the information flow of the learned adjacency matrix $\mathbf{A}$ and the intermediate node embedding matrix $\mathbf{Z}$ during the iterative procedure.
The presence of the blue arrow from $\mathbf{X}$ to $\mathbf{A}^{(0)}$ depends on whether there is a graph structure in the initial data.
The pseudocode of our methods is summarized in Algorithm \ref{BAGE}.

\begin{algorithm}  
	\caption{Algorithm to our methods (BAGE and VBAGE)}  
	\label{BAGE}  
	\begin{algorithmic} [1]
		\REQUIRE  Node features matrix $\mathbf{X}$, maximum iteration number $T$, parameters $\lambda$, $\alpha$ and $\tau$.\\
		\quad
		
		\textbf{Initialization:}\\
		\STATE \textbf{if} {There is the initial graph structure in the data:} 
			\STATE \quad Input $\mathbf{G} = \{\mathbf{V, E, X}\}$\\
				
		\STATE \textbf{else}: 
			\STATE \quad Initialize the adjacency matrix with Eq. (\ref{a_ij})
		\STATE \textbf{end}\\
		\quad 
		
		\textbf{Optimization:}
		\\
		 \STATE \textbf{for} epoch = 1,2,3, $\dots$, $T$ \textbf{do}:	
		 	\STATE \quad \textbf{if} epoch = 1,2,3, $\dots$, $\tau$:
				 \STATE \qquad  Update the adjacency matrix by Eqs. (\ref{a_ij}) and (\ref{A_alpha})
			\STATE \quad \textbf{end}
			\STATE \quad Compute the loss of BAGE by Eq. (\ref{loss_BAGE})
			\STATE \quad Compute the loss of VBAGE by Eq. (\ref{loss_VBAGE})\\	
			\STATE \quad Backpropagate loss and update $\mathbf{W}$\\				
		\STATE \textbf{end}	
		\ENSURE  Latent representation $\mathbf{Z}$\\			
	\end{algorithmic} 
\end{algorithm}

\section{Experiments}
We report our results on three tasks: node clustering, node classification, and graph visualization.
The benchmark datasets used in this paper include two real-world graph datasets and six general datasets.
The results not only demonstrate the advantage of our methods but also support the effectiveness of the adaptive learning of the adjacency matrix.

\subsection{Datasets} 

\noindent\textbf{Graph Datasets.}
In the node clustering task, two graph datasets Cora and Citeseer are used.
In order to verify the effectiveness of adaptive learning in our method, we perform incomplete processing on the adjacency matrix on these two graph datasets.
Incomplete processing means that the elements in the adjacency matrix are randomly set to 0 with a certain probability (Missing Ratio).
The detailed information of the two graph datasets is shown in Table \ref{graph dataset}.

\noindent\textbf{General Datasets.}
In the node classification task, six general datasets are used.
The is no graph structure in these datasets where we initialize the adjacency matrix by $k$NN for other methods which are based on the graph convolutional neural network.
The purpose of using these datasets is to verify the superiority of the adaptive learning of the adjacency matrix in our methods. 
The detailed information of the six general datasets is shown in Table \ref{general dataset}.

\begin{table*}[h]
	\caption{Information of the two graph datasets.}
	\centering
	\renewcommand
	\arraystretch{1.3}
	\begin{tabular}{lccccc}
		\toprule[1.5pt]
		Dataset  & \# Nodes & \# Links & \# Content Words & \# Features & \# Missing Ratio                     \\
		Cora     & 2,708    & 5,429    & 3,880,564        & 1,433       & \{0\%, 5\%, 10\%, 15\%, 20\%, 25\%, 50\%\} \\
		Citeseer & 3,327    & 4,732    & 12,274,336       & 3,703       & \{0\%, 5\%, 10\%, 15\%, 20\%, 25\%, 50\%\} \\
		\bottomrule[1.5pt]
	\end{tabular}
	\label{graph dataset}
\end{table*}

\begin{table}[h]
	\caption{Information of the six general datasets.}
	\centering
	\renewcommand
	\arraystretch{1.3}
	\begin{tabular}{l c c c c c c c}
		\toprule[1.5pt]		
		Datasets    & IMM   & ATT  & UMIST & COIL  & USPS & PIE  \\ \midrule[1pt]
		\# Nodes    & 240   & 400  & 575   & 1,440 & 2007 & 3332 \\ \midrule[1pt]
		\# Features & 1,024 & 1024 & 1024  & 1,024 & 256  & 4096 \\ 		
		\bottomrule[1.5pt]
	\end{tabular}
	\label{general dataset}
\end{table}

\subsection{Competitors}
We compare our algorithms against several state-of-the-art algorithms:

$\bullet$ \textbf{LGAE and LVGAE} \cite{salha2019keep}: is the latest improvement to GAE and VGAE, which replace the GCN encoder by a simple linear model w.r.t. the adjacency matrix of the graph.

$\bullet$ \textbf{ARGE and ARVGE} \cite{pan2018adversarially}: is the adversarial graph embedding framework that enforces latent representations to match a prior distribution, which is achieved by an adversarial training mode.

$\bullet$ \textbf{GAE and VGE} \cite{kipf2016variational}: is the classical graph autoencoder and variational graph autoencoder. And it is the first method to to apply graph convolution in the autoencoder.

$\bullet$ \textbf{SDNE} \cite{wang2016structural}: exploits the first order and second order proximity jointly to preserve the network structure and we use it as a baseline.

$\bullet$ \textbf{Spectral Clustering} \cite{ng2002spectral}: is a famous graph based clustering method and we use it as a baseline.

$\bullet$ \textbf{$k$-means} is the base of many clustering methods. Here we run $k$-means on raw node features as a baseline.

\subsection{Task 1: Node Clustering}

Like most papers based on the graph convolutional neural network, we apply our methods and other competitors on the node clustering task to verify the effect of graph embedding. \\

\noindent\textbf{Parameter Settings and Study.}
For the Cora and Citeseer datasets, we train all autoencoder-related models for 200 iterations and optimize them with the Adam algorithm.
The learning rate for the BAGE is 0.0001 and VBAGE is 0.001.
When the adjacency matrix is incomplete, the scale of adaptive learning $\alpha$ is 10\% and the number of adaptive learning iterations ($\tau$) is limited to 10-15.
The $\lambda$ is set as 0.01, and the elements of weight matrix $\beta$ is set as 20.
What's more, the parameters for other competitors are set to the values that make the best experimental results.

As for the impact of parameters, we study the influence of parameters $\lambda$ and $\beta$ on the experimental results.
The results and analysis of the parameter study are merged into the next subsection.
\\

\noindent\textbf{Metrics.}
To verify the effect of graph embedding, we run $k$-means on the learned latent representations to perform the node clustering task.
We run $k$-means 10 times with different initializations and report the mean and standard deviation of all methods.
To validate the clustering results, we employ two metrics: Accuracy (ACC) and Normalized Mutual Information (NMI).
\\

\noindent \textbf{Experimental Results and Analysis.}
The details of the experimental results on the node clustering are given in Tables \ref{Cora} and \ref{Citeseer}, where the best and second results have been highlighted.
Since the performance of competitors SDNE, Spectral Clustering (SC), and $k$-means is not very good, the results of those baselines are listed in a separate Table \ref{baselines}.
The results observations are as follows:

1) Our methods outperform all other competitors on the Cora dataset regardless of whether the adjacency matrix is incomplete.

2) In the Cora dataset, our methods (BAGE and VBAGE) are nearly 4\%-6\% higher than the second place (ARGE and ARVGE) when the missing ratio is little, and 7\%-9\% higher than the second place (ARGE and ARVGE) when the missing ratio is large.

3) In the Citeseer dataset, our methods perform 3\%-6\% higher than the other competitors when the missing ratio is less than 50\%.

\begin{table*}[]
	\caption{The node clustering results on the Cora dataset.}
	\centering
	\renewcommand
	\arraystretch{1.3}
	\begin{tabular}{lcccccccc}
		\toprule[1.5pt]		
		Methods                & Metrics & 0\%            & 5\%            & 10\%           & 15\%           & 20\%           & 25\%           & 50\%           \\ \midrule[1pt]
		\multirow{2}{*}{GAE}   & ACC(\%) & 59.53$\pm$2.00 & 58.27$\pm$3.68 & 55.13$\pm$2.74 & 57.19$\pm$2.30 & 54.57$\pm$1.91 & 55.58$\pm$3.60 & 48.77$\pm$2.73 \\
		& NMI(\%) & 40.48$\pm$1.09 & 38.82$\pm$1.78 & 32.64$\pm$1.12 & 36.67$\pm$1.77 & 35.10$\pm$1.61 & 36.98$\pm$1.69 & 25.23$\pm$1.60 \\
		\multirow{2}{*}{VGAE}  & ACC(\%) & 58.58$\pm$4.02 & 57.64$\pm$2.82 & 56.48$\pm$3.84 & 56.64$\pm$2.31 & 56.88$\pm$2.92 & 56.95$\pm$4.56 & 54.75$\pm$4.31 \\
		& NMI(\%) & 38.46$\pm$2.22 & 37.39$\pm$1.59 & 38.14$\pm$2.72 & 35.42$\pm$1.35 & 36.80$\pm$2.44 & 35.05$\pm$3.26 & 33.00$\pm$2.91 \\ \midrule[1pt]
		\multirow{2}{*}{ARGE}  & ACC(\%) & 68.64$\pm$0.60 & 67.66$\pm$0.69 & 65.73$\pm$1.62 & 63.37$\pm$2.54 & 60.69$\pm$1.20 & 59.44$\pm$1.78 & 55.21$\pm$0.52 \\
		& NMI(\%) & 49.22$\pm$0.17 & 46.30$\pm$0.91 & 45.12$\pm$1.22 & 44.07$\pm$1.72 & 42.07$\pm$0.30 & 39.52$\pm$0.83 & 34.33$\pm$0.42 \\
		\multirow{2}{*}{ARVGE} & ACC(\%) & 67.66$\pm$0.26 & 66.60$\pm$0.80 & 65.99$\pm$1.12 & 62.44$\pm$2.31 & 58.47$\pm$0.91 & 60.70$\pm$0.08 & 56.32$\pm$1.33 \\
		& NMI(\%) & 48.17$\pm$0.21 & 47.24$\pm$0.42 & 47.20$\pm$0.67 & 44.07$\pm$0.65 & 41.75$\pm$0.09 & 42.27$\pm$0.14 & 33.92$\pm$0.71 \\  \midrule[1pt]
		\multirow{2}{*}{LGAE}  & ACC(\%) & 58.42$\pm$0.90 & 57.46$\pm$3.93 & 57.82$\pm$1.64 & 53.02$\pm$2.47 & 51.55$\pm$2.81 & 54.66$\pm$2.86 & 45.12$\pm$2.35 \\
		& NMI(\%) & 36.63$\pm$0.49 & 39.71$\pm$1.39 & 39.34$\pm$0.84 & 36.62$\pm$1.44 & 32.89$\pm$0.92 & 35.46$\pm$0.94 & 23.84$\pm$1.28 \\
		\multirow{2}{*}{LVGAE} & ACC(\%) & 57.10$\pm$2.29 & 56.25$\pm$2.75 & 57.82$\pm$2.93 & 54.08$\pm$3.87 & 55.00$\pm$2.84 & 55.92$\pm$1.86 & 49.68$\pm$1.63 \\
		& NMI(\%) & 32.77$\pm$1.63 & 30.71$\pm$1.10 & 32.08$\pm$1.23 & 30.26$\pm$2.25 & 29.41$\pm$1.42 & 29.93$\pm$0.77 & 23.50$\pm$0.68 \\  \midrule[1pt]
		\multirow{2}{*}{BAGE}  & ACC(\%) & \textbf{72.57$\pm$0.68} & \textbf{71.21$\pm$0.18} & \textbf{70.32$\pm$0.33} & \textbf{72.74$\pm$0.12} & \textbf{69.43$\pm$2.27} & \textbf{66.59$\pm$0.87} & \textbf{64.22$\pm$0.29} \\
		& NMI(\%) & \textbf{56.91$\pm$0.49} & \textbf{54.73$\pm$0.15} & \textbf{52.39$\pm$0.17} & \textbf{52.43$\pm$0.24} & \textbf{49.96$\pm$0.79} & \textbf{51.88$\pm$0.54} & \textbf{48.53$\pm$0.25} \\
		\multirow{2}{*}{VBAGE} & ACC(\%) & \textbf{73.11$\pm$0.25} & \textbf{71.97$\pm$0.05} & \textbf{71.45$\pm$0.87} & \textbf{71.23$\pm$0.19} & \textbf{70.72$\pm$0.27} & \textbf{68.25$\pm$2.42} & \textbf{64.86$\pm$1.18} \\
		& NMI(\%) & \textbf{55.66$\pm$0.14} & \textbf{54.13$\pm$0.23} & \textbf{53.15$\pm$0.28} & \textbf{51.12$\pm$0.24} & \textbf{51.02$\pm$0.32} & \textbf{49.87$\pm$1.29} & \textbf{44.85$\pm$0.21}\\
	    \bottomrule[1.5pt]
	\end{tabular}
	\label{Cora}
\end{table*}

\begin{table*}[]
	\caption{The node clustering results on the Citeseer dataset.}
	\centering
	\renewcommand
	\arraystretch{1.3}
	\begin{tabular}{lcccccccc}
		\toprule[1.5pt]	
		Methods                & Metrics & 0\%                     & 5\%                     & 10\%                    & 15\%                    & 20\%                    & 25\%                    & 50\%                    \\ \midrule[1pt]
		\multirow{2}{*}{GAE}   & ACC(\%) & 54.54$\pm$3.45          & 54.19$\pm$2.62          & 50.95$\pm$2.27          & 49.17$\pm$1.50          & 50.96$\pm$1.76          & 44.25$\pm$2.76          & 43.51$\pm$2.81          \\
		& NMI(\%) & 27.04$\pm$1.53          & 24.40$\pm$2.12          & 21.49$\pm$1.44          & 20.55$\pm$1.09          & 22.33$\pm$1.28          & 16.25$\pm$1.62          & 16.18$\pm$1.73          \\
		\multirow{2}{*}{VGAE}  & ACC(\%) & 53.96$\pm$0.98          & 53.73$\pm$1.43          & 52.84$\pm$1.27          & 50.24$\pm$0.88          & 51.78$\pm$1.28          & 47.93$\pm$2.50          & 41.87$\pm$2.81          \\
		& NMI(\%) & 23.36$\pm$0.82          & 23.87$\pm$0.96          & 22.66$\pm$0.89          & 21.48$\pm$0.97          & 22.71$\pm$0.93          & 18.89$\pm$1.55          & 14.29$\pm$1.35          \\ \midrule[1pt]
		\multirow{2}{*}{ARGE}  & ACC(\%) & 58.67$\pm$1.20          & 57.95$\pm$0.62          & 54.80$\pm$1.19          & 50.49$\pm$1.68          & 51.66$\pm$0.99          & 51.40$\pm$1.42          & \textbf{45.42$\pm$0.14} \\
		& NMI(\%) & 31.33$\pm$0.85          & 30.69$\pm$0.41          & 27.75$\pm$1.18          & 24.17$\pm$1.08          & 24.81$\pm$0.55          & 24.18$\pm$0.82          & \textbf{20.26$\pm$0.26} \\
		\multirow{2}{*}{ARVGE} & ACC(\%) & 49.85$\pm$0.60          & 51.43$\pm$0.50          & 49.93$\pm$0.10          & 48.14$\pm$0.11          & 50.04$\pm$0.20          & 52.14$\pm$0.88          & 42.96$\pm$1.88          \\
		& NMI(\%) & 24.67$\pm$0.42          & 25.10$\pm$0.48          & 24.15$\pm$0.27          & 22.71$\pm$0.28          & 25.42$\pm$0.20          & 25.58$\pm$0.43          & 17.28$\pm$1.41          \\ \midrule[1pt]
		\multirow{2}{*}{LGAE}  & ACC(\%) & 56.66$\pm$0.96          & 56.93$\pm$0.88          & 56.58$\pm$0.94          & 55.08$\pm$1.14          & 55.63$\pm$1.01          & 54.74$\pm$0.42          & \textbf{46.83$\pm$1.00} \\
		& NMI(\%) & 27.22$\pm$0.59          & 28.41$\pm$0.73          & 27.00$\pm$0.86          & 26.67$\pm$0.88          & 26.38$\pm$0.91          & 25.21$\pm$0.34          & \textbf{28.48$\pm$0.54} \\
		\multirow{2}{*}{LVGAE} & ACC(\%) & 50.86$\pm$1.05          & 50.83$\pm$1.27          & 50.06$\pm$0.69          & 49.07$\pm$1.09          & 49.91$\pm$1.57          & 48.82$\pm$0.82          & 28.97$\pm$0.41          \\
		& NMI(\%) & 22.55$\pm$0.48          & 21.38$\pm$0.41          & 22.03$\pm$0.55          & 20.19$\pm$0.38          & 20.85$\pm$0.64          & 19.45$\pm$0.36          & 10.31$\pm$0.06          \\ \midrule[1pt]
		\multirow{2}{*}{BAGE}  & ACC(\%) & \textbf{64.64$\pm$0.19} & \textbf{62.19$\pm$0.34} & \textbf{58.83$\pm$2.23} & \textbf{57.96$\pm$1.14} & \textbf{59.64$\pm$0.48} & \textbf{55.18$\pm$2.12} & 39.91$\pm$3.57          \\
		& NMI(\%) & \textbf{39.07$\pm$0.30} & \textbf{37.99$\pm$0.21} & \textbf{36.03$\pm$1.13} & \textbf{32.13$\pm$0.61} & \textbf{35.17$\pm$0.35} & \textbf{26.17$\pm$0.98} & 14.62$\pm$1.85          \\
		\multirow{2}{*}{VBAGE} & ACC(\%) & \textbf{63.13$\pm$0.16} & \textbf{62.73$\pm$0.28} & \textbf{60.91$\pm$0.19} & \textbf{58.40$\pm$0.62} & \textbf{58.80$\pm$0.41} & \textbf{55.87$\pm$2.11} & 35.75$\pm$1.54          \\
		& NMI(\%) & \textbf{36.92$\pm$0.12} & \textbf{37.62$\pm$0.15} & \textbf{35.81$\pm$0.16} & \textbf{33.79$\pm$0.26} & \textbf{27.81$\pm$0.22} & \textbf{26.89$\pm$0.82} & 13.11$\pm$0.42    \\
		\bottomrule[1.5pt]     
	\end{tabular}
	\label{Citeseer}
\end{table*}

\begin{figure*}[h]
	\centering
	\setlength{\belowcaptionskip}{0pt}
	\subfigure [BAGE]
	{
		\includegraphics[height=3cm,width=3.2cm]{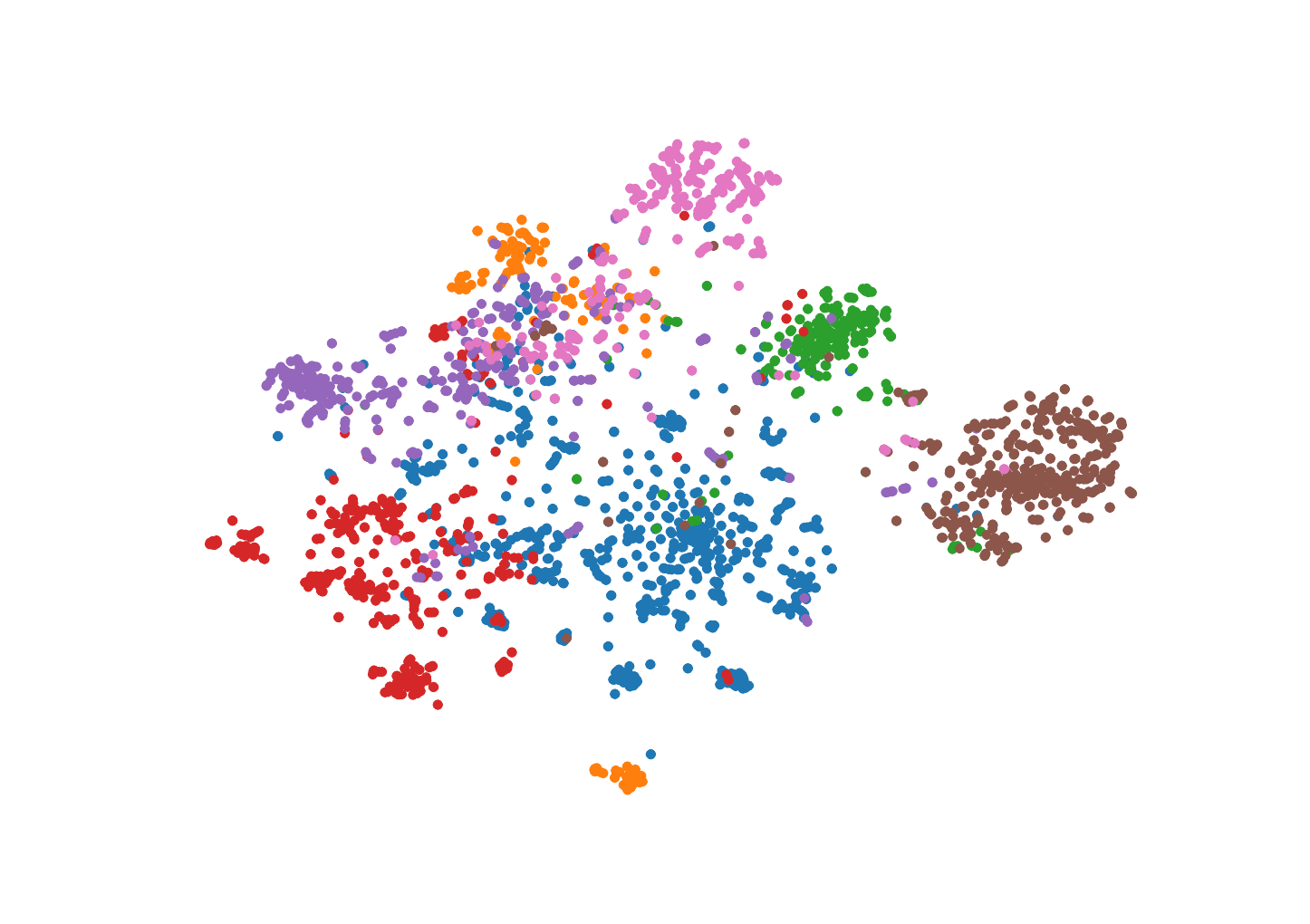}
	}
	\subfigure [VBAGE]
	{
		\includegraphics[height=3cm,width=3.2cm]{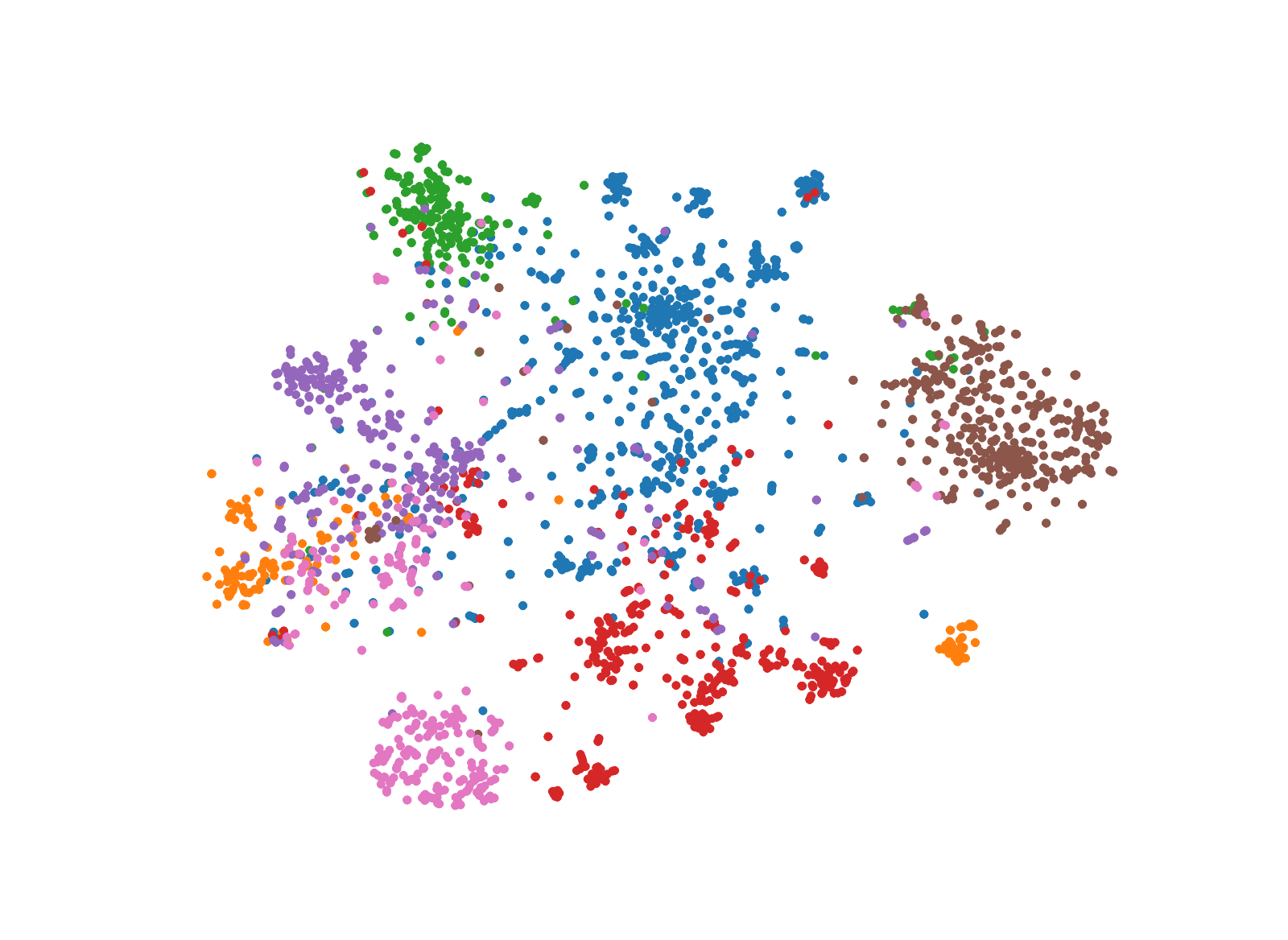}
	}
	\subfigure [ARGE]
	{
		\includegraphics[height=3cm,width=3.2cm]{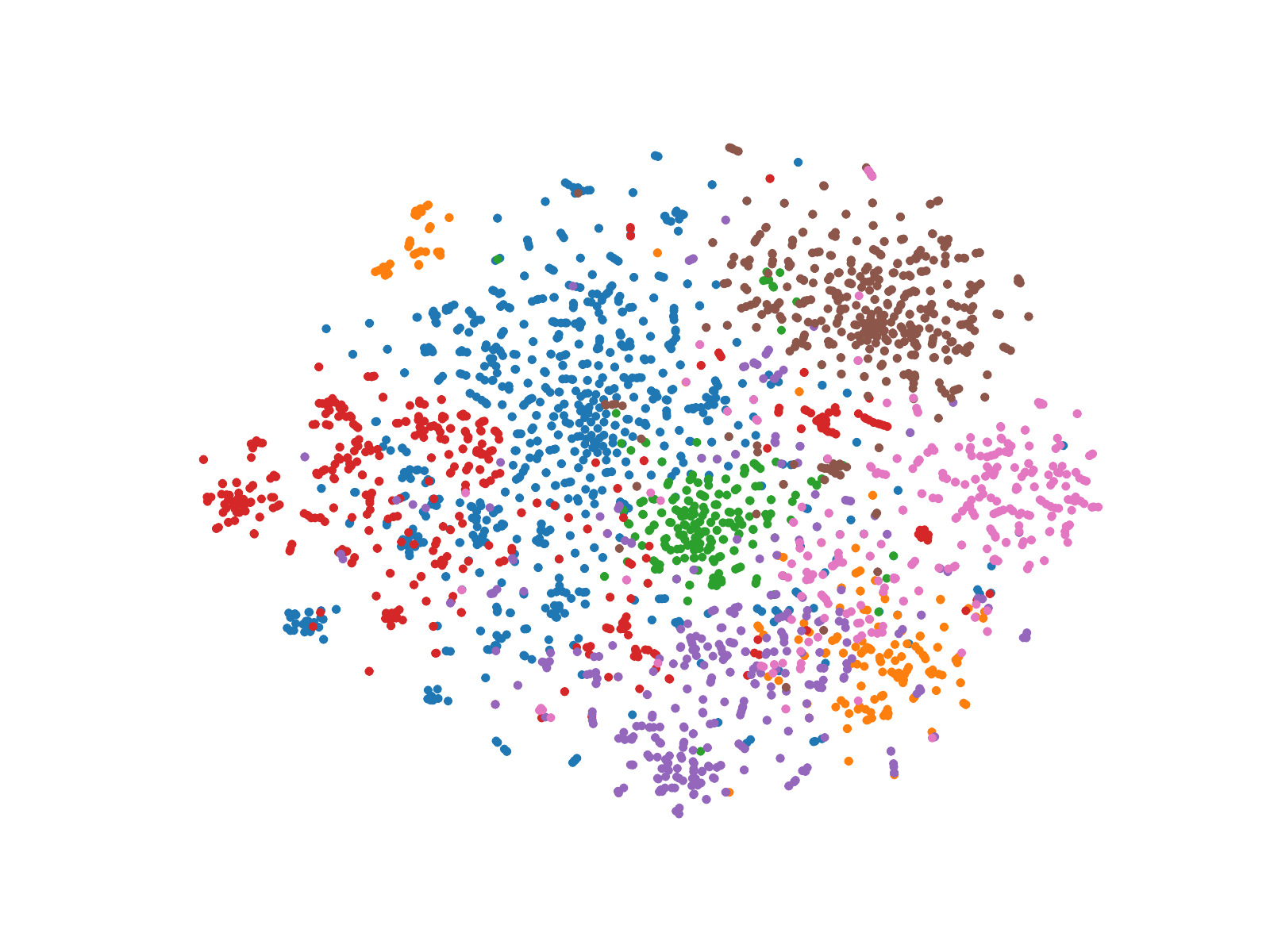}
	}
	\subfigure [GAE]
	{
		\includegraphics[height=3cm,width=3.2cm]{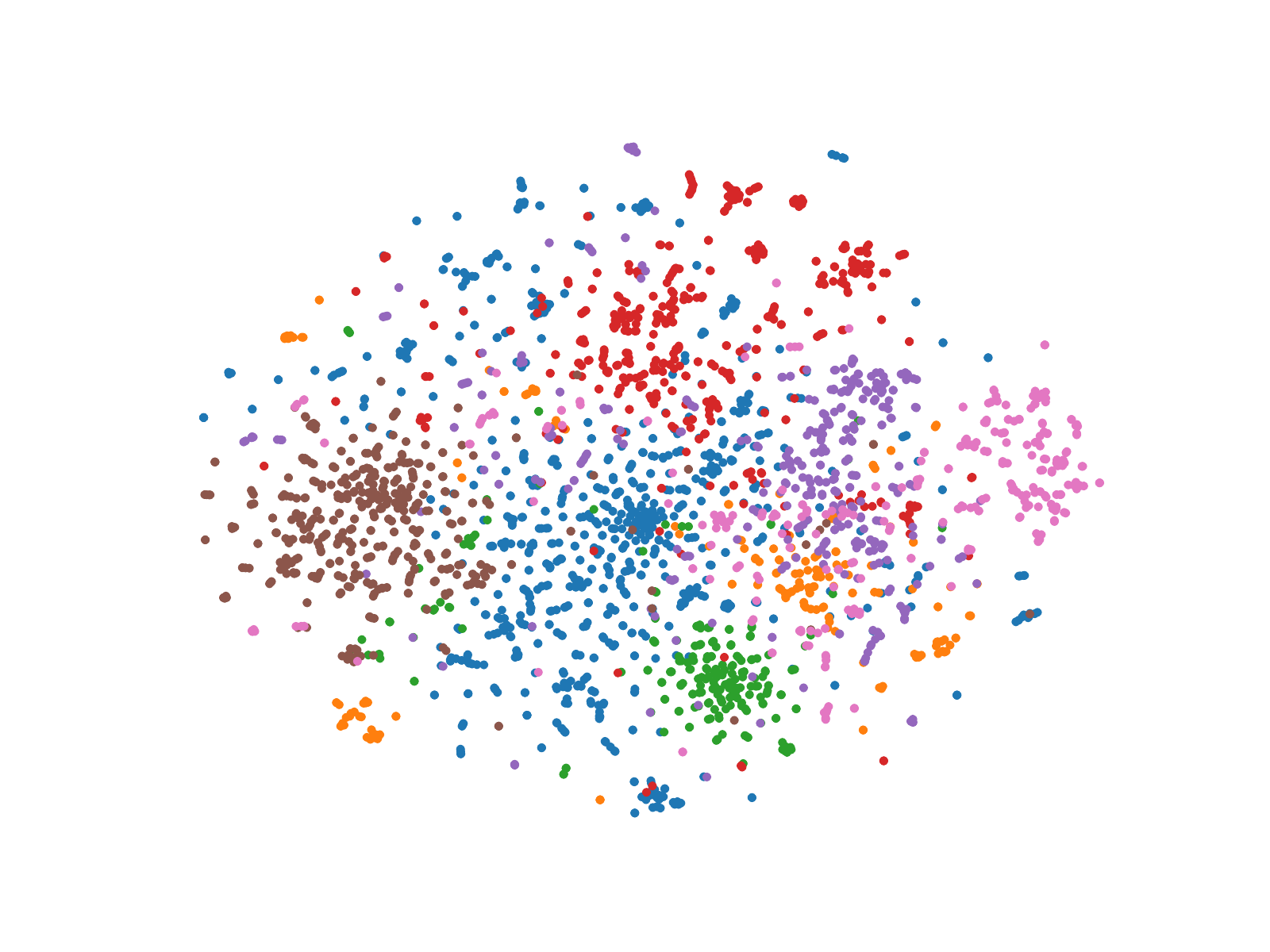}
	}
	\subfigure [LGAE]
	{
		\includegraphics[height=3cm,width=3.2cm]{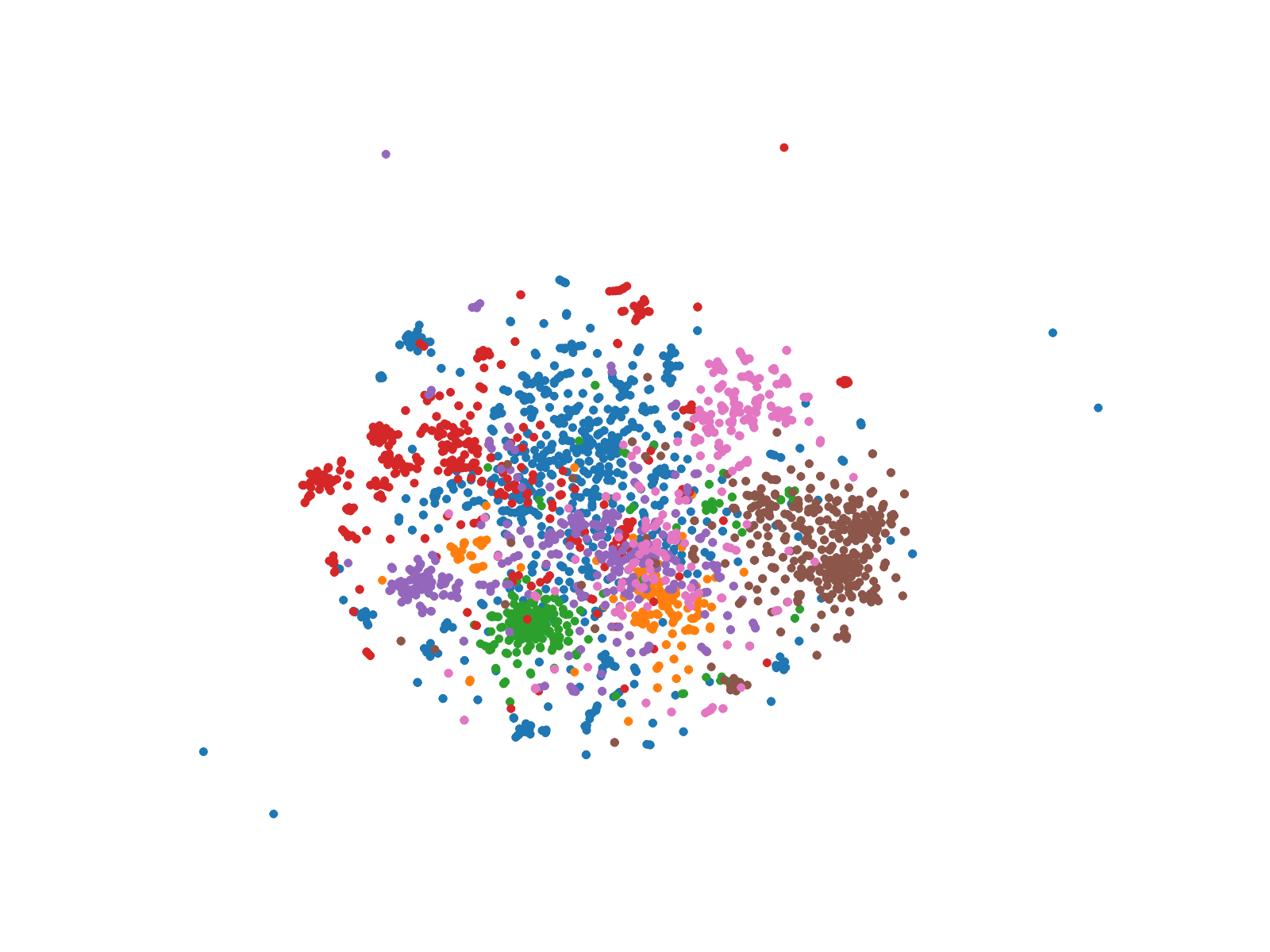}
	}
	\caption{The data visualization comparison on Cora.}	
	\label{SNE_Cora}
\end{figure*}

\begin{figure*}[]
	\centering
	\setlength{\belowcaptionskip}{0pt}
	\subfigure [BAGE]
	{
		\includegraphics[height=3cm,width=3.2cm]{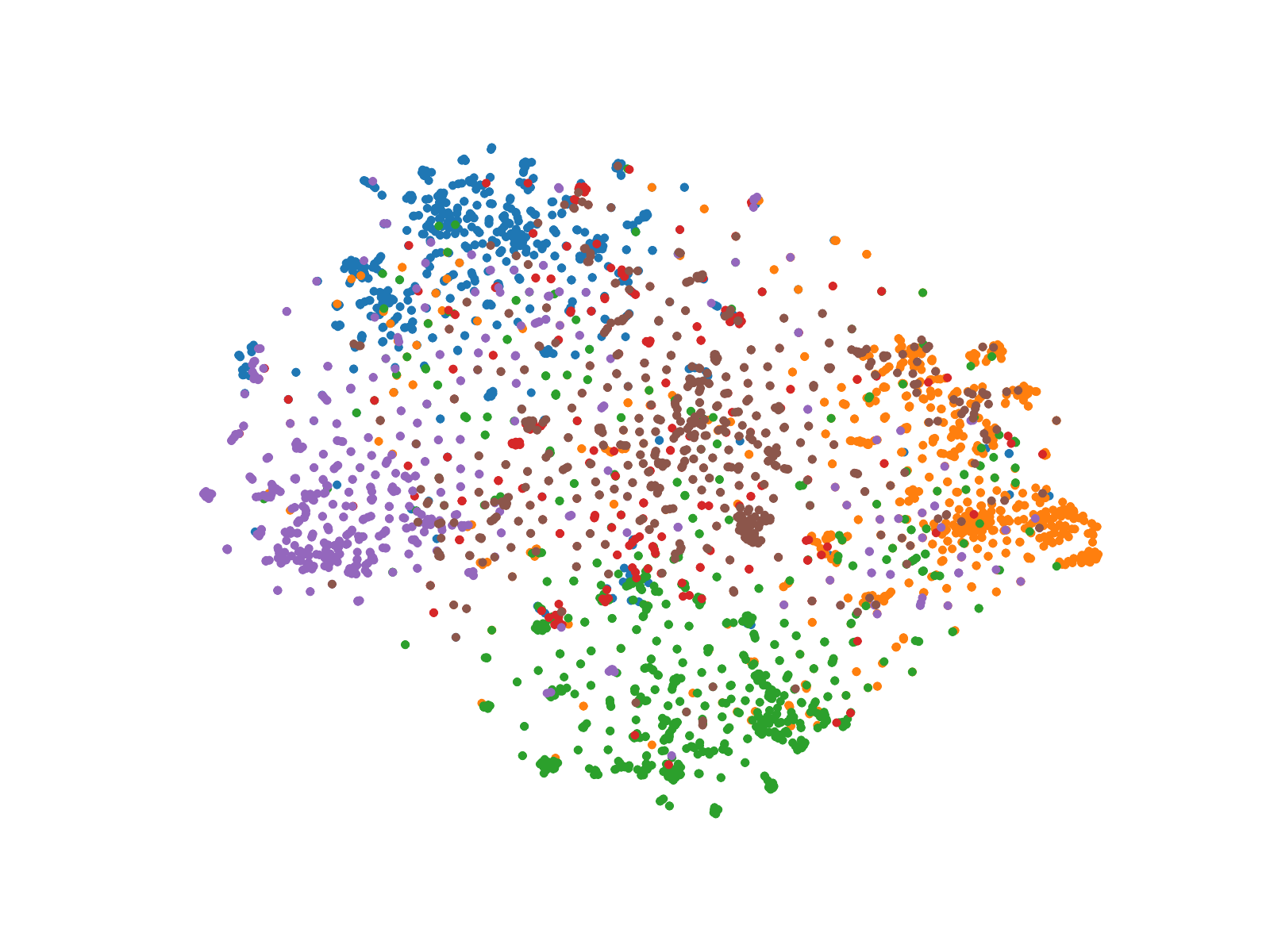}
	}
	\subfigure [VBAGE]
	{
		\includegraphics[height=3cm,width=3.2cm]{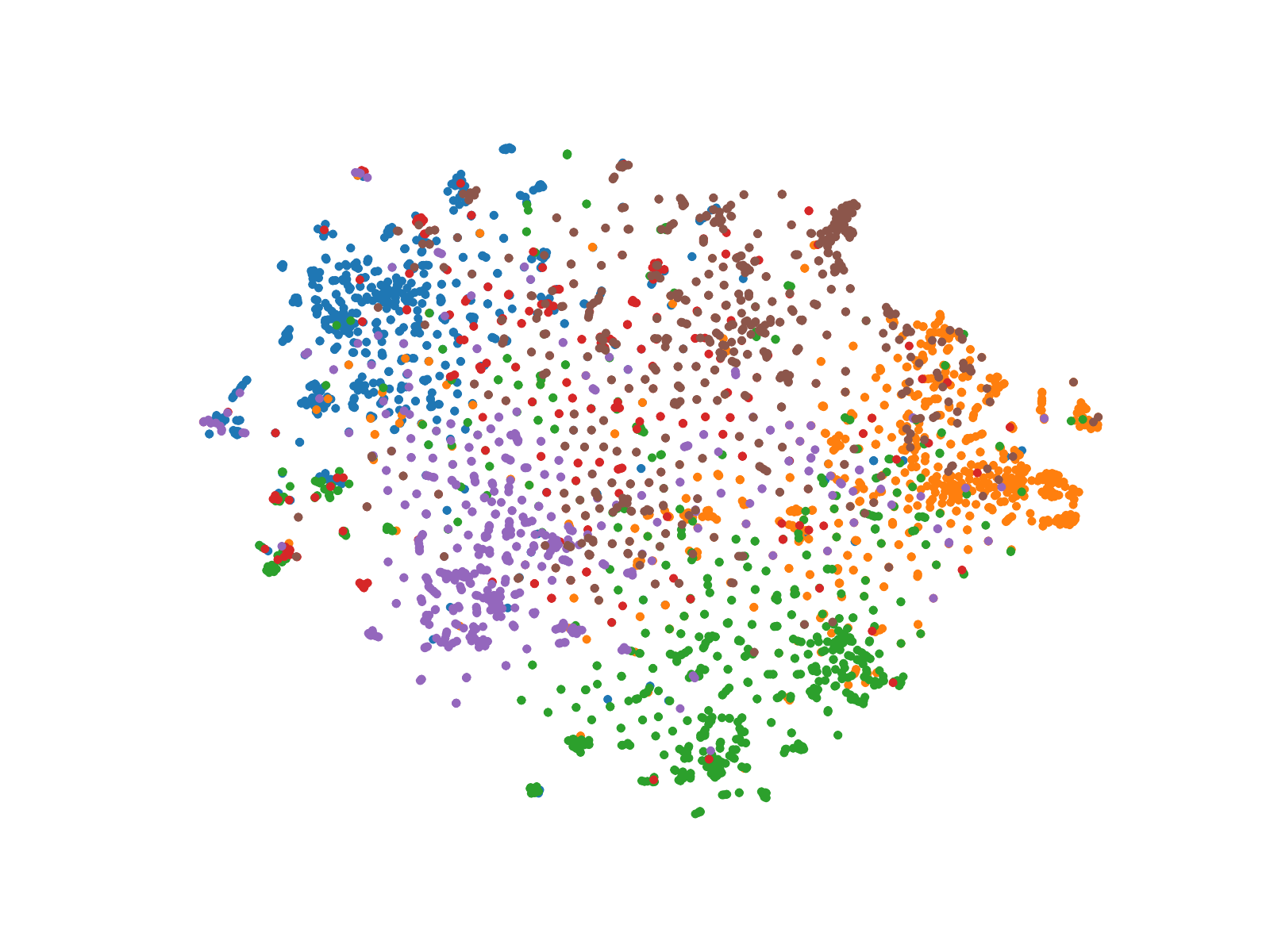}
	}
	\subfigure [ARGE]
	{
		\includegraphics[height=3cm,width=3.2cm]{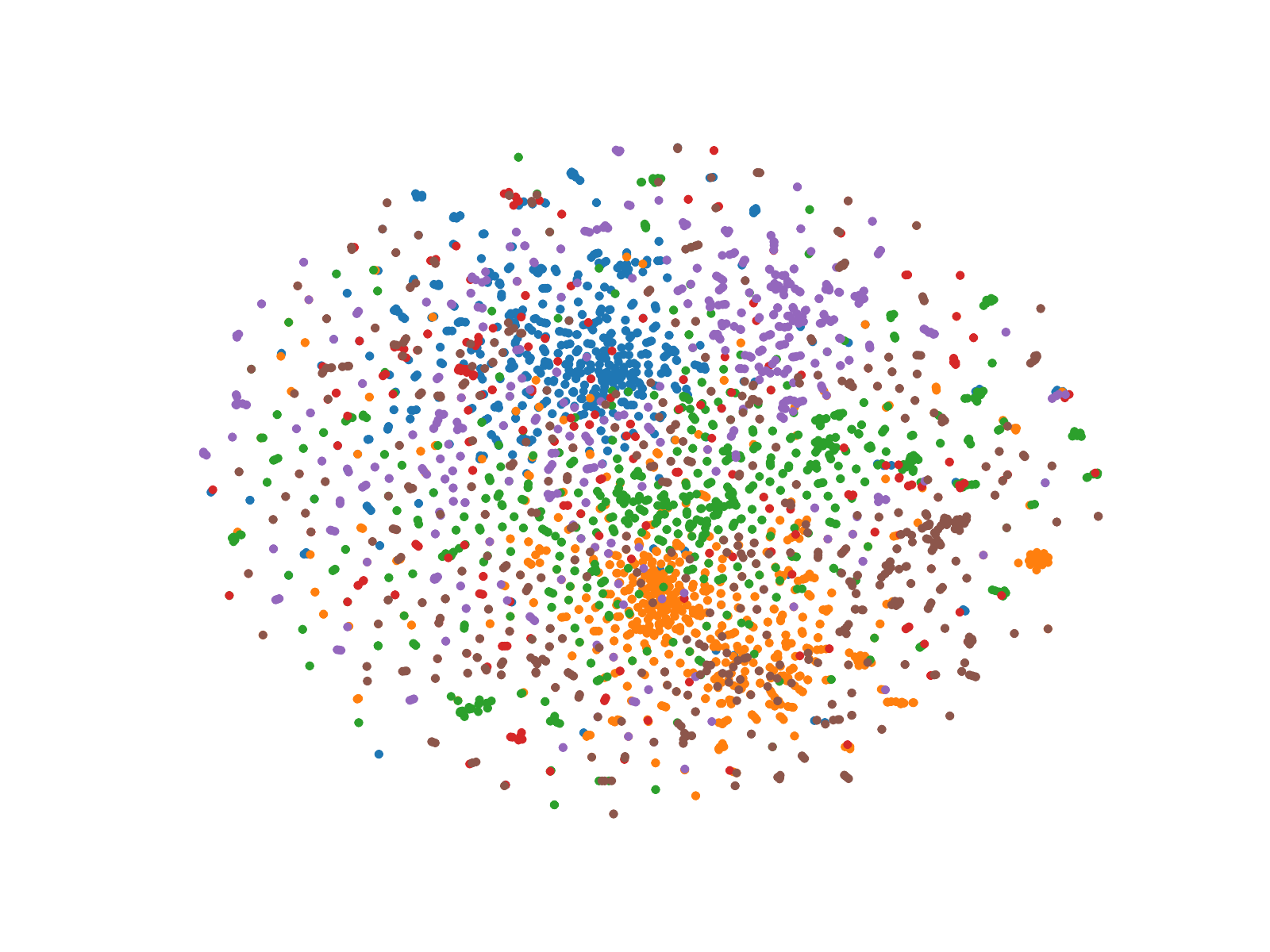}
	}
	\subfigure [GAE]
	{
		\includegraphics[height=3cm,width=3.2cm]{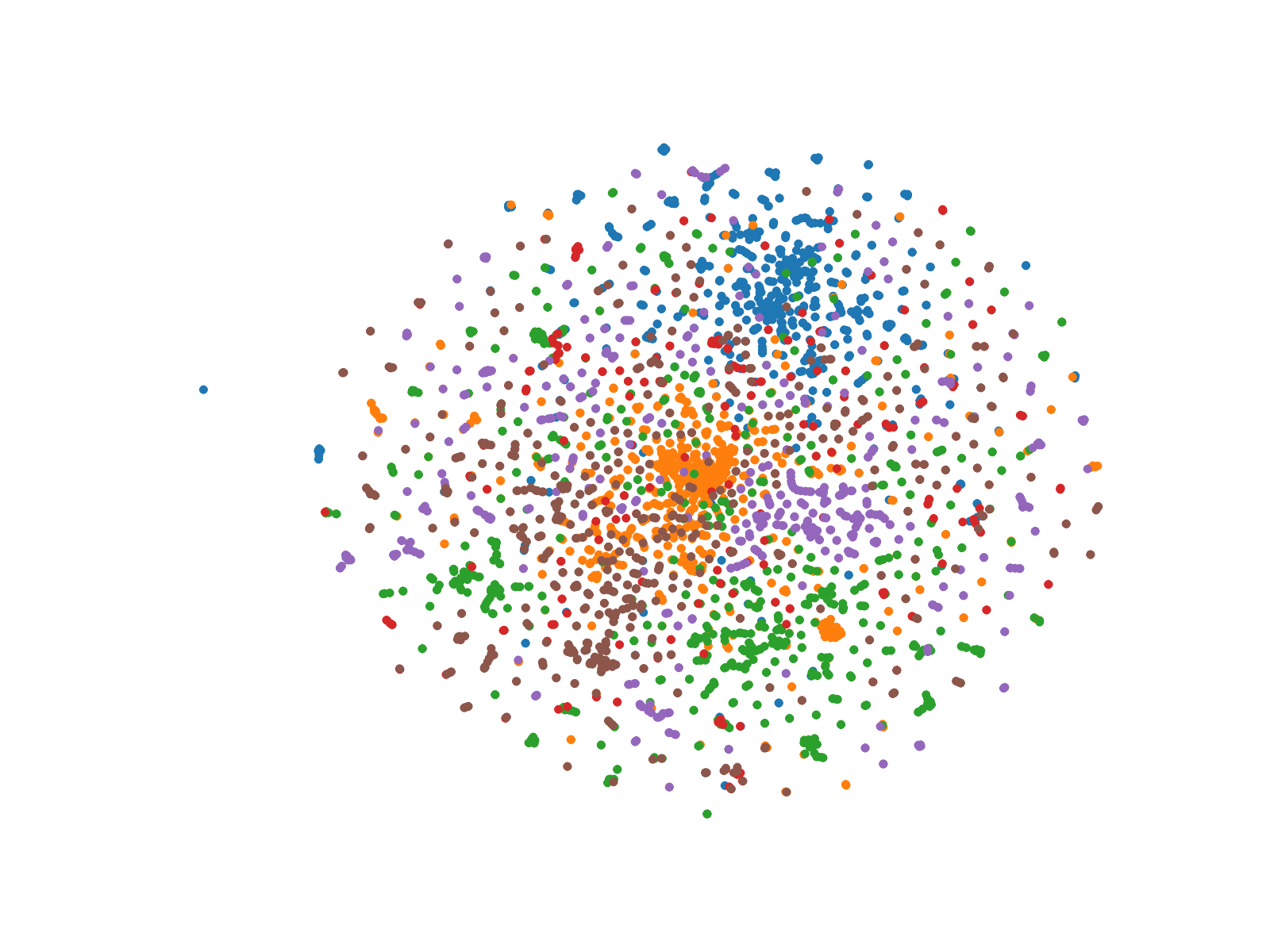}
	}
	\subfigure [LGAE]
	{
		\includegraphics[height=3cm,width=3.2cm]{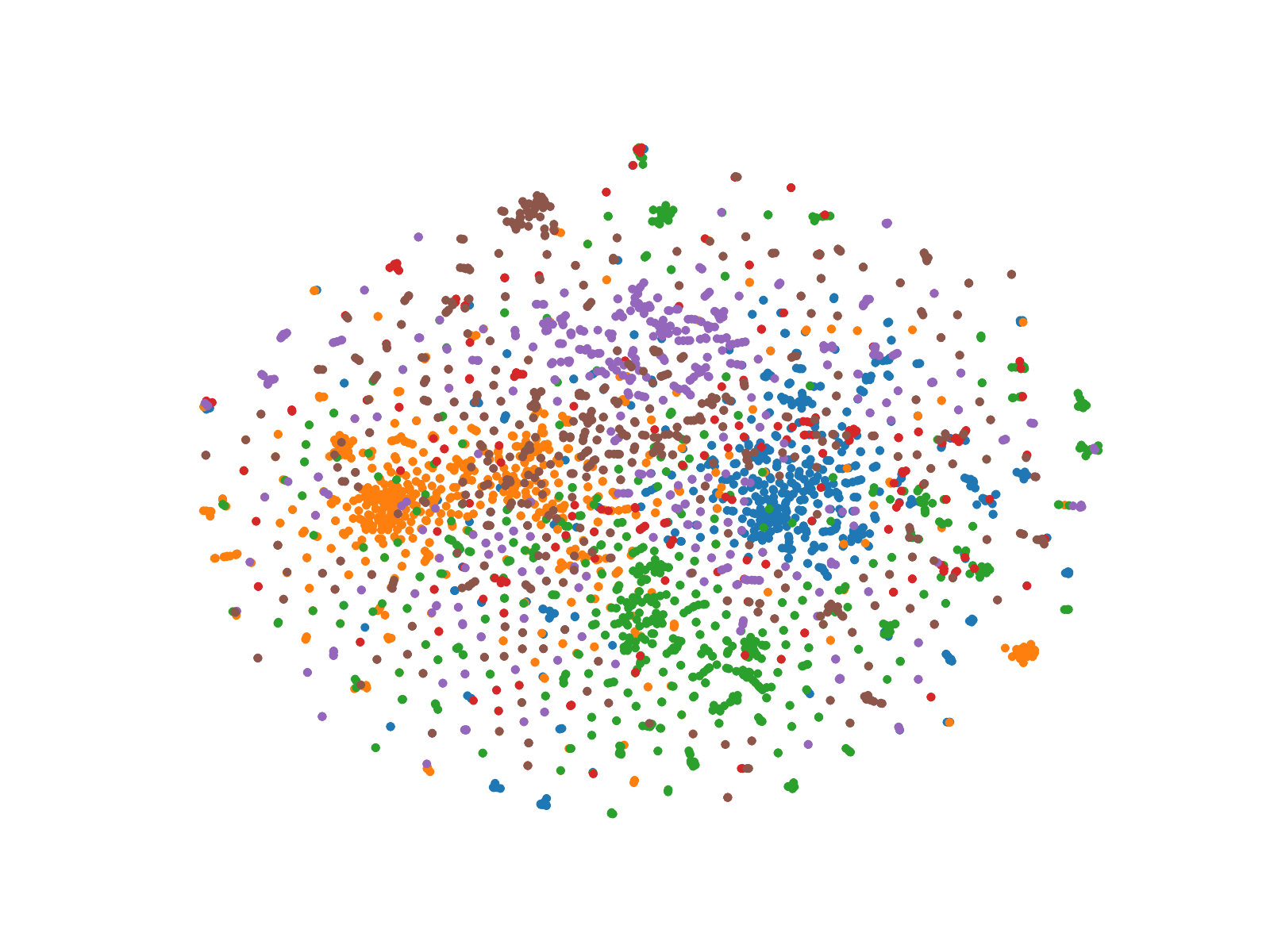}
	}
	
	\caption{The data visualization comparison on Citeseer.}
	\label{SNE_Citeseer}
\end{figure*}

\begin{table}[]
	\caption{The node clustering results for three baselines.}
	\centering
	\renewcommand
	\arraystretch{1.3}
	\begin{tabular}{lcccc}
		\toprule[1.5pt]	
		Methods                   & Metrics & SDNE           & SC                  & $k$-means      \\ \midrule[1pt]
		\multirow{2}{*}{Cora}     & ACC(\%) & 41.52$\pm$3.38 & 38.08$\pm$0.04      & 37.22$\pm$3.91 \\
		& NMI(\%) & 20.17$\pm$2.39 & 15.99$\pm$0.13      & 18.87$\pm$3.51 \\ \midrule[1pt]
		\multirow{2}{*}{Citeseer} & ACC(\%) & 30.21$\pm$0.67 & 21.46$\pm$0.00      & 43.80$\pm$5.83 \\
		& NMI(\%) & 4.44$\pm$0.33  & 1.72$\pm$0.00       & 20.63$\pm$4.67 \\
		\bottomrule[1.5pt]  
	\end{tabular}
	\label{baselines}
\end{table}

\subsection{Task 2: Node Classification}
In this task, we apply all methods on six general datasets to learn the latent representations which are then applied to the node classification task to verify the quality of the embedding results.
\\

\noindent\textbf{Parameter Settings and Study.}
All the parameter settings for our methods in this task are the same as in the node clustering task.
As for the parameters in other competitors, they are set as the values that make the best experimental results.

\begin{figure}[]
	\centering
	\setlength{\belowcaptionskip}{0pt}
	\subfigure
	{
		\includegraphics[height=3.5cm,width=4cm]{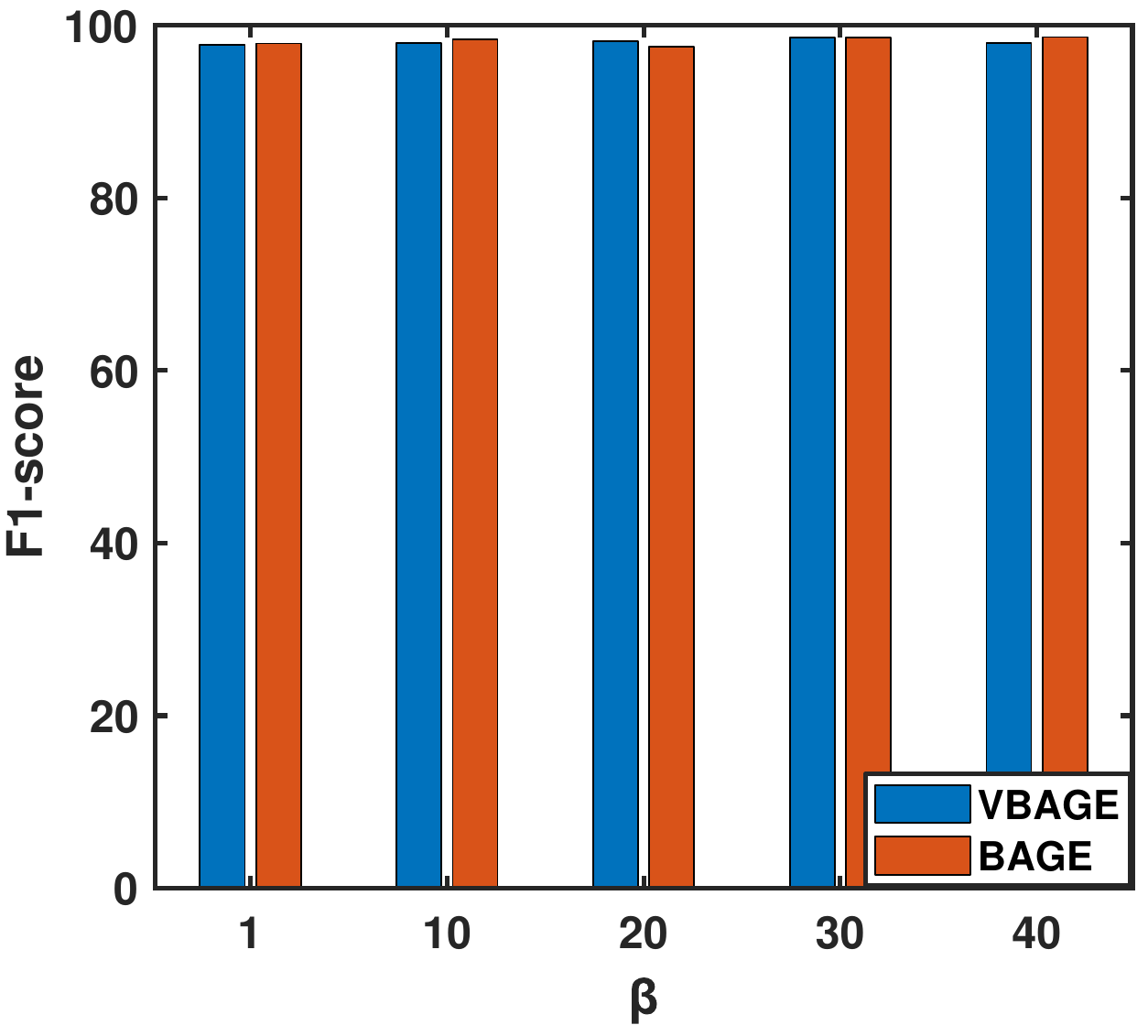}
	}
	\subfigure
	{
		\includegraphics[height=3.5cm,width=4cm]{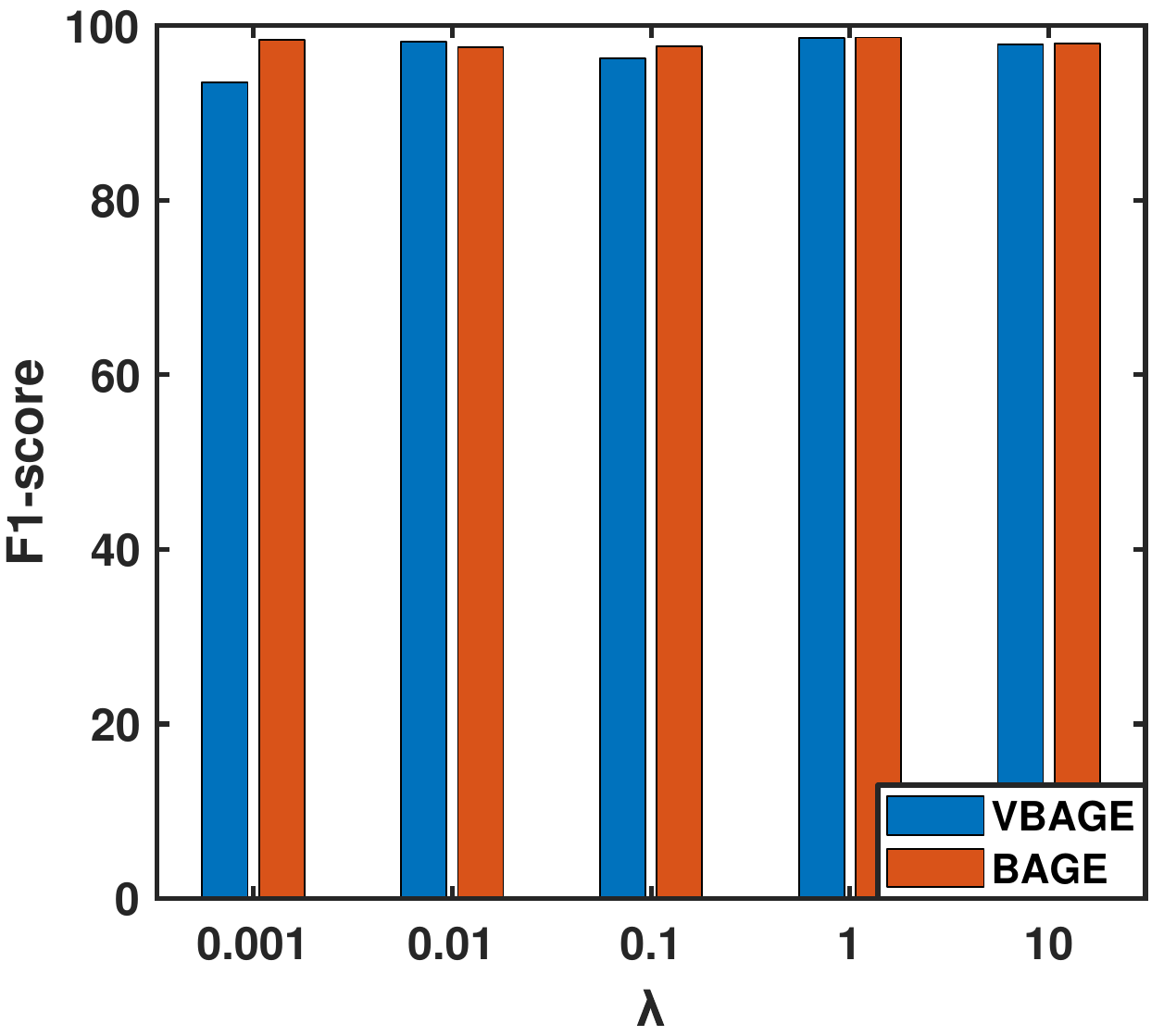}
	}
	\caption{The parameters study results on COIL dataset.}
	\label{parameters}
\end{figure}

For the impact of the parameters, we study the influence of parameters $\lambda$ and $\beta$ on the experimental results.
The value range of $\lambda$ is \{0.001, 0.01, 0.1, 1, 10\} and the value range of $\beta$ in weight matrix $\mathbf{B}$ is \{1, 10, 20, 30, 40\}.
The evaluation index is F1-score that is commonly used in the classification task.
Fig. \ref{parameters} shows the results of parameters study where the results show that the parameters' influence on our method on the node classification task is small.
\\

\noindent\textbf{Metrics.}
To verify the effect of graph embedding, we run SVM on the learned latent representations to perform the classification task.
The latent representations are divided into a training set (70\%) and a test set (30\%).
We first perform 10-fold cross-validation to select the best SVM model, and then we apply the selected SVM model on the test set.
To validate the node classification results, we employ the F1-score as the metric.
\\

\noindent \textbf{Experimental Results and Analysis.}
The details of the experimental results on the node classification task are displayed in Table \ref{F1} where the best and second results have been highlighted.
Due to space constraints, we only select a few outstanding competitors, i.e., ARGE, LGAE, GAE, and VGAE, to show in Table \ref{F1}.
The node classification results in Table \ref{F1} show the superiority of our method.

1) The performance of BAGE on the six general datasets has always remained in the top two and our method (BAGE and VBAGE) is higher than other methods 3\%-6\% in the COIL  dataset.

2) In UMIST and USPS, our methods (BAGE and VBAGE)  perform better than GAE, ARGE, and LGAE and the performance of our method occupies the top two.
 
3) In the PIE dataset, BAGE can achieve 95\% when other methods can only reach 91\%.

In summary, the experiment results in the node classification task fully illustrate the rationality and superiority of the adaptive learning of the adjacency matrix in our method.

\subsection{Task 3: Graph Visualization}
In this task, we apply our methods and several outstanding competitors to perform visualization in two graph datasets.
We visualize the Cora and Citeseer datasets in a two-dimensional space by applying the t-SNE \cite{van2014accelerating} algorithm on the learned embeddings. 

The results in Figs. \ref{SNE_Cora} and \ref{SNE_Citeseer} validate that by applying adaptive learning of the adjacency matrix, we can obtain a more meaningful layout of the graph data.

\begin{table}[]
	\caption{The F1-score (\%) on the node classification task.}
	\centering
	\renewcommand
	\arraystretch{1.3}
	\begin{tabular}{lcccccc}
		\toprule[1.5pt]	
		Methods & BAGE           & VBAGE          & ARGE           & LGAE           & GAE   & VGAE  \\ \midrule[1pt]
		COIL    & \textbf{97.50} & \textbf{98.15} & 94.33          & 93.85          & 93.86 & 92.84 \\ \midrule[1pt]
		ATT     & \textbf{86.59} & \textbf{87.54} & 83.87          & 83.50          & 78.48 & 76.24 \\ \midrule[1pt]
		IMM     & \textbf{36.71} & 25.81          & \textbf{32.39} & 30.05          & 20.43 & 19.54 \\ \midrule[1pt]
		UMIST   & \textbf{96.46} & \textbf{95.92} & 92.39          & 91.80          & 87.86 & 85.88 \\ \midrule[1pt]
		USPS    & \textbf{94.19} & \textbf{93.86} & 92.88          & 93.06          & 93.37 & 92.89 \\ \midrule[1pt]
		PIE     & \textbf{95.29} & 91.00          & 91.63          & \textbf{91.69} & 87.69 & 87.54 \\
		\bottomrule[1.5pt]  
	\end{tabular}
	\label{F1}
\end{table}

\section{Conclusion}
In this paper, we propose two novel unsupervised graph embedding methods, \emph{unsupervised graph embedding via adaptive graph learning} (BAGE) and \emph{unsupervised graph embedding via variational adaptive graph learning} (VBAGE).
Aiming at the problem that the existing GAE methods are sensitive to the adjacency matrix, we embed the adaptive learning to the framework, which enhances the robustness of the model.
In addition, the adaptive learning mechanism expands the application range of GAEs on graph embedding and is able to initialize the adjacency matrix without be affected by the parameter $k$.
Furthermore, the learned latent representations are embedded in the laplacian graph structure to preserve the topology structure of the graph in the vector space.
Experimental studies on several datasets demonstrated that our methods (BAGE and VBAGE) outperform baselines by a wide margin in node clustering, node classification, and graph visualization tasks.

\medskip
\small

\bibliographystyle{IEEEbib.bst}
\bibliography{BAGE.bib}

\end{document}